\documentclass[letterpaper, 10 pt, conference]{ieeeconf}  % Comment this line out if you need a4paper

\IEEEoverridecommandlockouts                              % This command is only needed if 
\usepackage{graphics} % for pdf, bitmapped graphics files
\usepackage{float}
\usepackage{stfloats}
\usepackage{amsmath}
\usepackage{amssymb}
\usepackage{multicol}
\usepackage{graphicx}
\usepackage{layouts}
\usepackage{wrapfig}
\usepackage{caption}
\usepackage{graphicx,subcaption}
\usepackage{easyReview}
\usepackage{printlen}
\usepackage{bm}
\usepackage{siunitx}
\usepackage{colortbl}	% to color table background
\usepackage[utf8]{inputenc}
\usepackage{pgf}
\usepackage{tikz}
\usepackage{tikzscale}
\usepackage{pgfplots}
\usepackage[hidelinks]{hyperref}
\usepackage{threeparttable}
\usepackage{tabularx}
\usepackage{makecell}
\usepackage{multirow}
\usepackage{pgf}

\pgfplotsset{compat=1.14}
\usepgfplotslibrary{external}
\usepgfplotslibrary[external]
\tikzset{external/force remake}

\newcommand{\gencoor}{\boldsymbol{q}}
\newcommand{\genvel}{\boldsymbol{v}}
\newcommand{\nonlinear}{\boldsymbol{h}}
\newcommand{\torque}{\boldsymbol{\tau}}
\newcommand{\force}{\boldsymbol{\lambda}}

\newcommand{\lmom}{\boldsymbol{l}_G}
\newcommand{\amom}{\boldsymbol{k}_G}
\newcommand{\com}{\boldsymbol{c}_G}
\newcommand{\lmomrate}{\dot{\boldsymbol{l}}_G}
\newcommand{\amomrate}{\dot{\boldsymbol{k}}_G}
\newcommand{\comrate}{\dot{\boldsymbol{c}}_G}

\title{\LARGE \bf
Centroidal State Estimation Based on the Koopman Embedding for Dynamic Legged Locomotion}
\author{
Shahram Khorshidi$^{*}$ \and Murad Dawood$^{*}$ \and Maren Bennewitz% <-this % stops a space
\thanks{$^{*}$ Equal contribution.}
\thanks{All authors are with the Humanoid Robots Lab, University of Bonn, Germany. Maren Bennewitz and Murad Dawood are additionally with the Lamarr Institute for Machine Learning and Artificial Intelligence, Bonn, Germany. This work has partially been funded by the Deutsche Forschungsgemeinschaft (DFG, German Research Foundation) under Germany's Excellence Strategy, EXC-2070 -- 390732324 -- PhenoRob and by the European Union’s Horizon Europe research and innovation programme under grant agreement No~101070405~(DigiForest) as well as by the Federal Ministry of Education and Research of Germany and the state of North Rhine-Westphalia as part of the Lamarr Institute for Machine Learning and Artificial Intelligence, Germany.}
}

\begin{document}

\maketitle
\thispagestyle{empty}
\pagestyle{empty}

%%%%%%%%%%%%%%%%%%%%%%%%%%%%%%%%%%%%%%%%%%%%%%%%%%%%%%%%%%%%%%%%%%%%%%%%%%%%%%%%
\begin{abstract}
In this paper, we introduce a novel approach to centroidal state estimation, which 
plays a crucial role in predictive model-based control strategies for dynamic legged locomotion. Our approach uses the Koopman operator theory to transform the robot's complex nonlinear dynamics into a linear system, by employing dynamic mode decomposition and deep learning for model construction. We evaluate both models on their linearization accuracy and capability to capture both fast and slow dynamic system responses. We then select the most suitable model for estimation purposes, and integrate it within a moving horizon estimator. This estimator is formulated as a convex quadratic program to facilitate robust, real-time centroidal state estimation. Through extensive simulation experiments on a quadruped robot executing various dynamic gaits, our data-driven framework outperforms conventional Extended Kalman Filtering technique based on nonlinear dynamics. Our estimator addresses challenges posed by force/torque measurement noise in highly dynamic motions and accurately recovers the centroidal states, demonstrating the adaptability and effectiveness of the Koopman-based linear representation for complex locomotive behaviors. Importantly, our model based on dynamic mode decomposition, trained with two locomotion patterns (trot and jump), successfully estimates the centroidal states for a different motion (bound) without retraining.
\end{abstract}

%%%%%%%%%%%%%%%%%%%%%%%%%%%%%%%%%%%%%%%%%%%%%%%%%%%%%%%%%%%%%%%%%%%%%%%%%%%%%%%%
\section{Introduction}
State estimation plays a crucial role in the deployment of predictive state feedback controllers for dynamic locomotion in legged robots. The underactuation, hybrid nature of contact dynamics, and nonlinear behavior of these robots pose substantial challenges for achieving accurate state estimation. This issue is particularly significant for model-based controllers that depend on precise centroidal state information, including the Center of Mass (CoM) position and momentum, to maintain stability and maneuverability during dynamic tasks. However, the dynamic nature of legged locomotion, characterized by rapid, intermittent contacts with various surfaces, introduces significant modeling uncertainties and measurement noise, particularly in the Force/Torque (F/T) data obtained from robot end-effectors, complicating the state estimation process.
\begin{figure}[!t]
	\centering
	\includegraphics[width=0.88\linewidth]{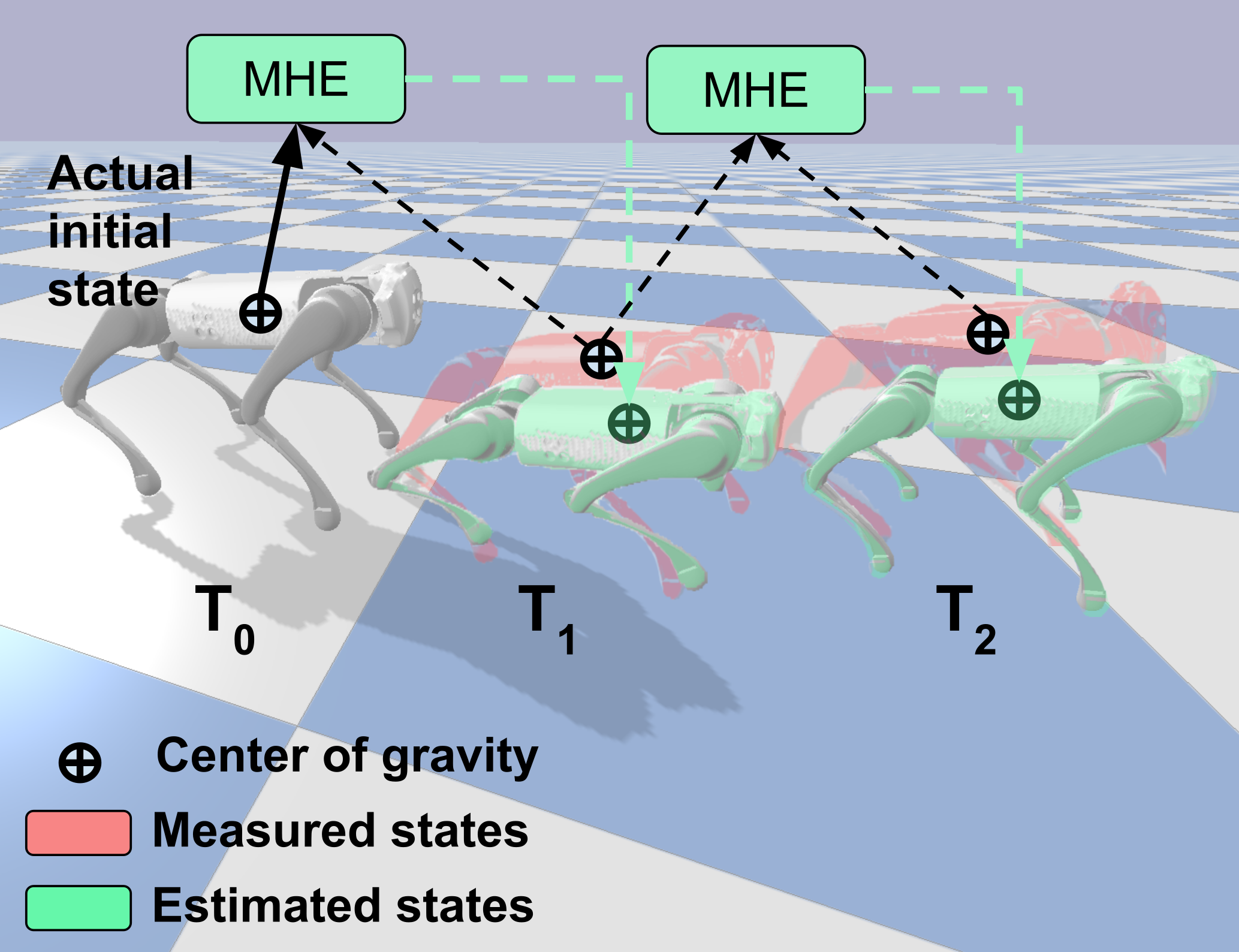}
	\caption{Moving horizon estimator utilizes the recent history of state measurements and system inputs, to predict the optimal states.}
	\vspace{-7mm}
	\label{fig:pipeline}
\end{figure}

Traditionally, the Extended Kalman Filter (EKF) has been used for centroidal state estimation using the contact wrenches with nonlinear centroidal dynamics \cite{rotella2015humanoid}. Yet, EKF struggles with dynamic motions where intermittent contact forces can greatly amplify F/T sensor noise, leading to reduced estimation accuracy. Additionally, model-based kinematic methods for computing centroidal states are prone to noise in joint and base velocity measurements. Although low-pass filters can mitigate this noise, they introduce latency that may destabilize high-frequency controllers.

Given the complexity and nonlinearity of legged robot dynamics, exploring linear representations and approximations of these dynamics that align with established theories in linear control and state estimation offers significant advantages. The Koopman operator theory represents a promising direction, by mapping the state space to an infinite-dimensional vector space where dynamics evolve linearly~\cite{Koopman1931}. Although deriving such a mapping remains challenging, recent advances in data-driven model identification algorithms, such as Dynamic Mode Decomposition (DMD)~\cite{Schmid2008dmd}, have gained popularity for their strong connection with the Koopman operator theory \cite{Mezic2005, Mezic2013}. This perspective is valuable for dynamic legged locomotion, as it circumvents the simplifying assumptions of conventional models, potentially enhancing control and estimation techniques for more general locomotive patterns.

In this study, we introduce a novel approach to centroidal state estimation for legged robots, leveraging the Koopman operator theory and embedding the robot's nonlinear dynamics into a linear system. This linearization is crucial as it allows for the formulation of the Moving Horizon Estimator~(MHE) as a convex Quadratic Program (QP). MHE utilizes the recent history of state measurements and system inputs, shown in Fig. \ref{fig:pipeline}, to produce optimal state estimates, thus effectively mitigating the challenges posed by F/T sensor noise and the computational delays of filtering noisy kinematic measurements. Our approach offers a more resilient estimation framework in highly dynamic motions and is less susceptible to the drawbacks of conventional EKF methods.

To our knowledge, this is the first application of the Koopman embedding for centroidal state estimation in legged robots. By considering the centroidal dynamics of legged robots, our framework's generality allows for its application across a broad spectrum of legged robot platforms with dynamic motions.

The main contributions of this work are as follows:
\begin{itemize}
	\item We develop and apply the Koopman embedding to the nonlinear centroidal dynamics of legged robots, using DMD and deep learning methods. 
	\item We evaluate the learned models in terms of the linearization error and the nature of captured dynamics, to select the most suitable model for estimation purposes.
	\item We formulate an MHE framework based on the linear DMD model for real-time state estimation.
\end{itemize}
Finally, through simulation experiments, we demonstrate the effectiveness of our approach across various dynamic gaits on a quadruped robot, and its superior performance compared against a model-based filtering estimation technique.
\section{Related Work}
Research in legged robot state estimation can broadly be categorized into two main areas: base and centroidal state estimation. Predominantly, recent studies have focused on estimating the base state, a task that presents notable challenges in floating base systems as the base pose cannot be directly measured. The EKF has emerged as a prevalent choice for this task due to its balance of simplicity, efficiency, and effectiveness.

In base state estimation, a common approach involves fusing Inertial Measurement Unit (IMU) data with leg odometry to estimate the base position, velocity, and orientation \cite{bloesch2013state}, \cite{rotella2014state}, \cite{fallon2014drift}. To mitigate drift in unobservable base positions and yaw angle, while simultaneously facilitating environmental mapping, additional sensor modalities such as LiDAR and cameras are frequently integrated \cite{fallon2014drift}, \cite{camurri2020pronto}, \cite{Dhedin2022visulainertial}. Moreover, more advanced and sophisticated methodologies leveraging factor graphs \cite{fourmy2021contact} and invariant Kalman filters \cite{hartley2020contact,ramadoss2021} have been introduced to enhance state estimation process. 

For centroidal state estimation, F/T measurements from the end-effectors of legged robots provide valuable information for enhancing the estimation of these states \cite{rotella2015humanoid}, managing contact switch events \cite{fourmy2021contact, rotella2018contact, bledt2018contact}, and improving overall state estimation accuracy \cite{geoff2020, valsecchi2020quadrupedal}. Some robot designs prioritize low leg inertia to achieve agile motions \cite{grimminger2020open}. To do that, they often omit F/T sensing, prompting alternative estimation strategies, such as leveraging joint torque measurements within an EKF framework for centroidal dynamics \cite{Khorshidi2023}. Moreover, optimization-based methods have been investigated for centroidal state estimation \cite{Bailly2021ddp}.

Emerging machine learning innovations are enhancing legged robot locomotion and state estimation, providing new capabilities and insights that complement and extend the performance of traditional methods. For instance, deep neural network architectures that learn displacement measurements from IMU data \cite{buchanan2021learning}, and estimate IMU bias terms for factor graph integration \cite{Buchanan2023bias}, have shown drift reduction for proprioceptive base state estimation in the case of impaired exteroceptive sensing. Furthermore, the separation of estimator and control policy learning for dynamic gaits has proven more effective than end-to-end learning \cite{Khadiv2023}, underscoring the importance of state estimation in both model-based and learning-based control frameworks in legged robot locomotion.

Data-driven algorithms based on the Koopman theory~\cite{Koopman1931}, and its approximation using DMD \cite{Schmid2008dmd}, offer novel insights into dynamical system modeling. DMD efficiently extracts low-order models from complex, high-dimensional systems and identifies spatial-temporal coherent modes from measurement data. Extensions of this method to incorporate the effect of control inputs has been proposed in \cite{Joshua2016dmdc}, capable of producing an input-output model. In addition, optimal
control strategy based on this representation is employed for a class of nonlinear dynamical systems \cite{Ian2019active}. Moreover, the subsequent development of extended DMD \cite{Williams2014edmd} for mapping the state space through nonlinear observables, while innovative, is hampered by its reliance on handcrafted heuristic functions, and introduces subjectivity and limitations in the captured dynamics.

Alternatively, deep learning techniques have the potentials to autonomously learn these complex mappings, eliminating the need for predefined heuristics. These methods, explored in various studies, employ an autoencoder structure to encode states into a latent space. Linearity is subsequently enforced on the latent variables to determine the state and input matrices through different methodologies. In \cite{lusch2018deep}, the authors enforced linearity on latent variables through a trainable linear layer that yields the state matrix. Conversely, \cite{takeishi2017learning, morton2018deep} addressed the challenge by solving a linear least-squares fit problem, resulting in the extraction of the state-transition matrix. Notably, \cite{Bounou2021onlinekoopman} demonstrated the necessity of regularly updating the state-transition matrix for improved capturing of system dynamics, showcasing enhanced performance compared to \cite{lusch2018deep, morton2018deep}.
\section{Fundamentals}
\subsection{Notation}
\begin{itemize}
    \item Throughout the paper, we use small letters to specify scalars and scalar-valued functions, bold small letters for vectors and vector-valued functions, and capital letters for matrices.
    \item $I_n \in \mathbb{R}^{n \times n}$ is the identity matrix.
    \item $\left\| A \right\|_{F}$ is the Frobenius norm of matrix $A$.
    \item $A^{\dagger}$ stands for the pseudo-inverse of matrix $A$.
	\item $\left\|\boldsymbol{x}\right\|_2^2$ denotes the squared Euclidean norm of vector $\boldsymbol{x}$.
	\item $\left\| \boldsymbol{x} \right\|_{P}$ represents the weighted Euclidean norm of vector~$\boldsymbol{x}$, such that $\left\| \boldsymbol{x} \right\|^2_{P} = \boldsymbol{x}^\top P\: \boldsymbol{x}$.
\end{itemize}
\subsection{Centroidal Dynamics}
The rigid body dynamics of a floating-base system can be written as 
\begin{equation}\label{eq:rigid_body}
    M(\gencoor) \dot \genvel + \nonlinear(\gencoor,\genvel) = S^\top \torque + J_c^\top \force,
\end{equation}
where $M \in \mathbb{R}^{(n+6) \times (n+6)}$ is the mass-inertia matrix, $\gencoor \in \mathit{SE}(3) \times \mathbb{R}^n$ denotes the configuration vector, $\genvel \in \mathbb{R}^{n+6}$ is the vector of generalized velocities, and $\nonlinear \in \mathbb{R}^{n+6}$ is a concatenation of nonlinear terms including centrifugal, Coriolis and gravitational effects. $S \in \mathbb{R}^{n \times (n+6)}$ is a selection matrix that separates the actuated and unactuated degrees of freedom, $\torque \in \mathbb{R}^n$ represents the vector of joint torques, $J_c \in \mathbb{R}^{6d \times (n+6)}$ is the Jacobian of $d$ feet in contact, and finally $\force \in \mathbb{R}^{6d}$ is the vector of contact wrenches.

These dynamics can be split into actuated and unactuated parts, where the unactuated dynamics is equivalent to the Newton-Euler equations of the center of mass (CoM) \cite{Wieber2006}, and by considering the relation between linear momentum and CoM linear velocity, they can be described as
\begin{align}\label{eq:centroidal_dynamics}
	\begin{bmatrix}
		{\comrate} \\
		{\lmomrate} \\
		{\amomrate}
	\end{bmatrix} &=     
	\begin{bmatrix}
	\frac{1}{m} \lmom \\	
	 m\boldsymbol{g} + \sum_{i=1}^{o} b_i \boldsymbol{f}_i \\ 
	\sum_{i=1}^{o} b_i\left( (\boldsymbol{r}_i - \boldsymbol{c}_G) \times \boldsymbol{f}_i + \boldsymbol{\tau}_i \right)
	\end{bmatrix},
\end{align}
where $\boldsymbol{c}_G \in \mathbb{R}^3$ represents the CoM position, and $\lmom\in \mathbb{R}^3$ and $\amom \in \mathbb{R}^3$ are linear and angular momentum w.r.t. the CoM, respectively. $m$ is the robot total mass, $\boldsymbol{g}$ is the gravitational acceleration vector, and $\boldsymbol{f}_i$, $\boldsymbol{\tau}_i$ and $\boldsymbol{r}_i$ account for the $i^{th}$ end-effector force, torque and location respectively. $b_i$ serves as a binary integer indicating whether the end-effector is in contact, and $o$ is the total number of end-effectors.

In this study, we focus on a quadruped robot and assume point-contact feet, leading to the exclusion of torque vectors at the end-effectors in our equations. Nonetheless, our findings are applicable to humanoid robots as well, provided that the full contact wrenches are incorporated into the learning process.
\section{Centroidal State Estimation}
\subsection{Problem Statement}
Our objective is to estimate the centroidal states of a legged robot, represented by $\boldsymbol{x} \in \mathbb{R}^9 = \begin{bmatrix}
\com, \lmom, \amom \end{bmatrix}^\top$. We utilize an approximated and learned Koopman embedding of the centroidal dynamics as the dynamic model in our MHE estimation process.

As demonstrated in \cite{rotella2015humanoid}, direct measurements of the centroidal states from generalized joint states, kinematics, and the robot's inertial properties are prone to significant noise and modeling inaccuracies. This necessitates the adoption of a state estimation framework to recover these quantities more accurately and without delay. Consequently, the measurement model is linear w.r.t. the states, described by
\begin{equation}\label{eq:measurement_model}
	\boldsymbol{y} = C \boldsymbol{x},
\end{equation}
where $\boldsymbol{y}$ is the measurement and $C = I_9$.
\subsection{Koopman Embedding}
The Koopman operator \cite{Koopman1931}, offers a transformative perspective on analyzing nonlinear dynamical systems by casting them within linear, albeit infinite-dimensional, systems. This operator acts on observable functions within the system's state space, linearly evolving these observables over time, independent of the system's nonlinear nature. Despite its infinite-dimensional characteristic, the Koopman operator is a powerful tool for dissecting complex dynamics into modes with distinct frequencies and growth rates, allowing linear analysis techniques to be applied to nonlinear systems. In our work, we employ two different methods to derive this approximation.
\subsubsection{Dynamic Mode Decomposition (DMD)}
The advent of numerical methods, especially DMD and its variants, has made computing approximations to the Koopman modes practical, broadening the theory's applicability to fields with complex or unknown dynamics. DMD aims to break down complex, high-dimensional data into modes, each tied to a specific frequency and growth/decay rate. These modes encapsulate the system's essential dynamics in a simplified yet dynamic-faithful representation. As an extension to this method, DMD with Control (DMDC) incorporates the effect of inputs to extract a low-order model while maintaining consistency with the underlying dynamics \cite{Joshua2016dmdc}. 

Our approach, inspired by the Koopman operator theory via DMDC approximation, aims to linearly represent centroidal dynamics as follows
\begin{align}\label{eq:linear_sys}
	\boldsymbol{x}_{k+1} \approx A\boldsymbol{x}_k + B\boldsymbol{u}_k,
\end{align}
where $\boldsymbol{x}_k$ and $\boldsymbol{u}_k$ denote state and input vectors at time step~$k$, respectively. To measure states, we incorporate kinematic information. Additionally, we consider the end effector force vectors and relative end-effector positions w.r.t. CoM position, as the vector of inputs into DMDC formulation.

Using $m$ samples obtained from robot motion, we structure the data matrices as
\begin{align}\label{eq:data_matrix}
	X &= [\boldsymbol{x}_1, \boldsymbol{x}_2, \dots , \boldsymbol{x}_{m-1}], \nonumber\\
	X^{'} &= [\boldsymbol{x}_2, \boldsymbol{x}_3, \dots , \boldsymbol{x}_{m}], \nonumber\\
	U &= [\boldsymbol{u}_1, \boldsymbol{u}_2, \dots , \boldsymbol{u}_{m-1}].
\end{align}
The data-driven nature of DMDC approach allows the integration of new trajectory data into the specified matrices, given that each trajectory's data matrices are structured as~\eqref{eq:data_matrix}.

Subsequently, the linear system representation in \eqref{eq:linear_sys} can be extended into matrix form to include these data matrices
\begin{align}\label{eq:matrix_linear_sys}
	X^{'} \approx A X + B U,
\end{align}
where DMDC solves the subsequent least-square problem to approximate mappings $A$ and $B$
\begin{align}\label{eq:least_square}
	\underset{
		\substack{
			K
		}
	}{\mathrm{argmin}}
	&\frac{1}{2}\left\|X^{'}-K\Omega\right\|_{F}^2,
\end{align}
where $K = [A \: B]$ and $\Omega = \left[ \begin{smallmatrix} X \\ U \end{smallmatrix} \right]$.

A common approach to address this problem employs singular value decomposition of $\Omega$, as per \cite{Joshua2016dmdc}
\begin{align}\label{eq:svd}
	K^{*} = X^{'} \Omega^{\dagger}.
\end{align}
This approach necessitates careful selection of a singular-value threshold to either dampen the solution or truncate $\Omega$'s spectrum.
\subsubsection{Deep Learning Koopman (DLK)}
\begin{figure}[!t]
	\centering
	\includegraphics[width=1.0\linewidth]{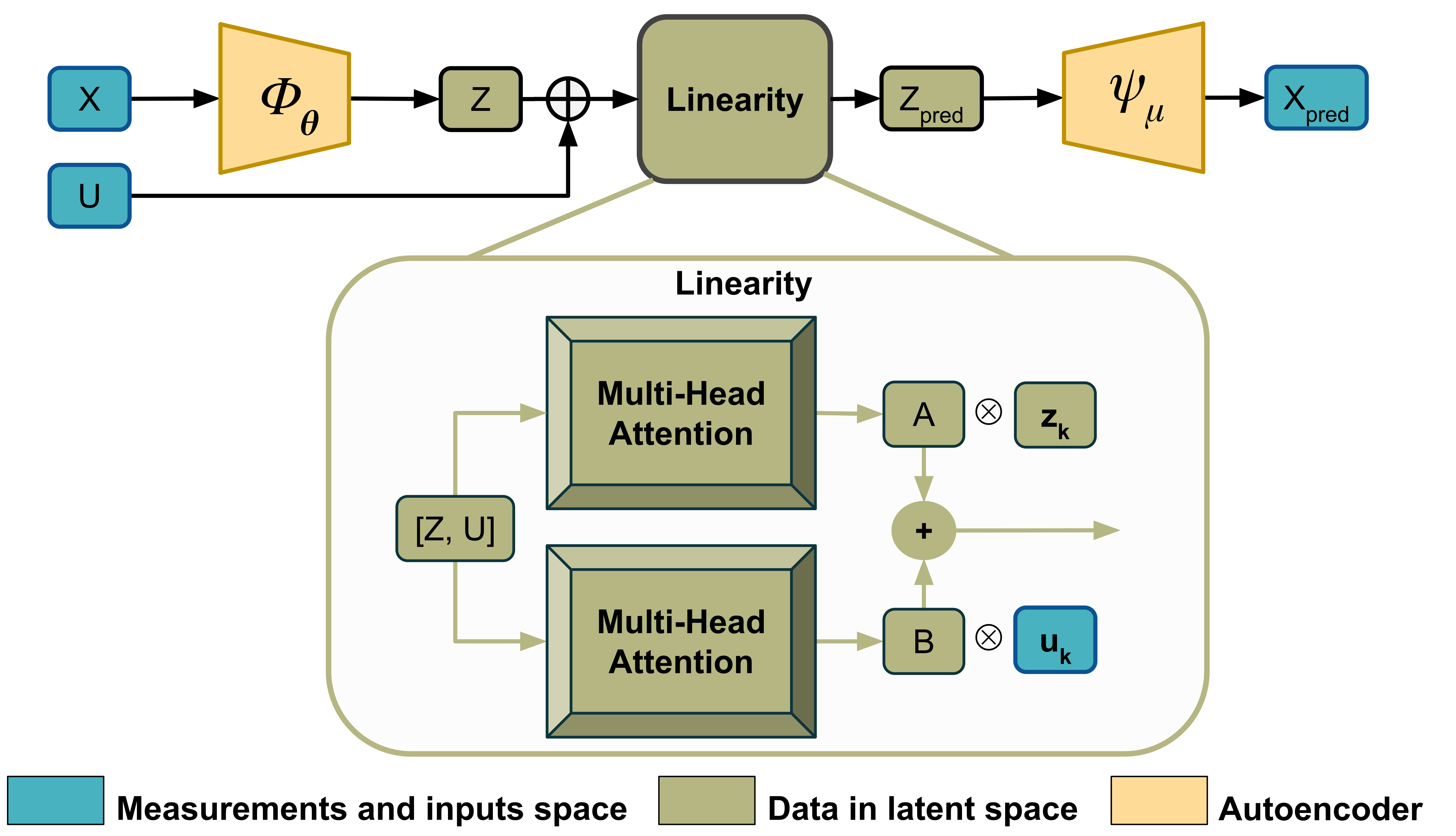}
	\caption{The neural network architecture to learn the Koopman embedding approximation of centroidal dynamics.}
	\vspace{-5mm}
	\label{fig:nn_architecture}
\end{figure}

Our objective is to acquire a latent space representation, denoted as $\boldsymbol{z}$, for states $\boldsymbol{x}$ using an encoder $\boldsymbol{\phi}_\theta$. Subsequently, we aim to learn a linear state representation $(A, B)$ of the dynamics within the latent space $\boldsymbol{z}$, governed by the recurrent relation similar to~\eqref{eq:linear_sys}. A schematic overview of the neural network architecture is presented in Fig. \ref{fig:nn_architecture}, where in the linearity block we enforce the linear representation \eqref{eq:linear_sys}. The latent state $\boldsymbol{z}_{k+1}$ is then decoded by the decoder $\boldsymbol{\psi}_\mu$ to obtain the predicted state $\hat{\boldsymbol{x}}_{k+1}$. Here $\theta$ and $\mu$ denote the weights of the encoder and decoder networks. 

To train the network, trajectories are partitioned into segments of $T$ steps, to account for the cyclic nature of the legged robot motion, which repeats approximately every two seconds. Recognizing the temporal dependencies in the trajectories, where each state evolves from the previous one and the input,  we propose a novel approach. We adopt multi-head attention \cite{vaswani2017attention} to effectively process and capture the system's dynamics in the latent space.

For each sample, we encode the first $j$ states into a latent space $Z$, and concatenate this matrix with the first $j$ inputs $U$. The resulting matrix, $[Z, U]$, is then processed through two multi-head attention models to derive the state matrix $A$ and the input matrix $B$. For the remaining $T-j$ states, we propagate the initially encoded state $\boldsymbol{z}_{j}$ and the remaining inputs through the linear dynamics as \eqref{eq:linear_sys}.

The loss functions for our model are defined as follows
\begin{subequations}\label{eq:loss_fn}
	\begin{align}
		L_{\text{Recon},\theta,\mu} &= \sum_{t=1}^{T}\left\|\boldsymbol{x}_{t}-\boldsymbol{\psi}_{\mu}(\boldsymbol{\phi}_{\theta}(\boldsymbol{x}_{t}))\right\|_2^2, \label{subeq:rec_loss}\\
		L_{\text{Pred},\theta,\mu} &= \!\!\!\sum_{t=j+1}^{T}\!\!
		\left\|\boldsymbol{x}_{t}-\boldsymbol{\psi}_{\mu}(A^{t-j}\boldsymbol{\phi}_{\theta}(\boldsymbol{x}_{j}) + B^{t-j}\boldsymbol{u}_{j})\right\|_2^2, \label{subeq:pred_loss}  \\
		L_{\text{Total}\theta,\mu} &= L_{\text{Recon},\theta,\mu} + L_{\text{Pred},\theta,\mu}. \label{subeq:loss_function}
	\end{align}
\end{subequations}
where \eqref{subeq:rec_loss} is the reconstruction loss, which trains the autoencoder to accurately encode and decode states into and from the latent space, and \eqref{subeq:pred_loss} is the prediction loss, ensuring the accuracy of linear predictions within the latent space. Subsequent experiments demonstrate that utilizing multi-head attention to represent $A$ and $B$ results in accurate predictions of future states without the need to frequently update the state and input matrices.
\subsection{Moving Horizon Estimation}
In our estimation approach, we have adopted an MHE framework for the state estimation process. To articulate the general formulation of this estimator, we begin with a nonlinear discrete-time stochastic system, described as follows
\begin{align}\label{eq:discrete_nonlinear_model}
	\boldsymbol{x}_{k+1} &= \boldsymbol{f}(\boldsymbol{x}_{k},
	\boldsymbol{u}_{k}) + \boldsymbol{w}_k,\nonumber\\
	\boldsymbol{y}_k &= \boldsymbol{h}(\boldsymbol{x}_k) + \boldsymbol{v}_k,
\end{align}
where $\boldsymbol{x}_k$, $\boldsymbol{y}_k$ and $\boldsymbol{u}_k$  represent the state, measurement, and input vectors at time step $k$, respectively. We model the disturbances in the system dynamics and measurement model as additive Gaussian noise, represented by $\boldsymbol{w}_k$ for the system equation and $\boldsymbol{v}_k$ for the measurement model.

The objective of MHE is to minimize the following cost function \cite{rao2001constrained}
\begin{subequations}\label{eq:mhe}
	\begin{align}
		\underset{
			\substack{
				{X}_{0:n+1},\\ {W}_{0:n}, {V}_{0:n}
			}
		}{\mathrm{min}}
		&\frac{1}{2}\left\|\boldsymbol{x}_0-\tilde{\boldsymbol{x}}_0\right\|_{P_x}^2 \!
		+\! \sum_{k=0}^{N-1} \left(\frac{1}{2}\|\boldsymbol{w}_k\|_{P_{w}}^2 \! + \frac{1}{2}\|\boldsymbol{v}_k\|_{P_{v}}^2\right), \label{subeq:cost_function} \\
		\mathrm{s.t.\; \;} 
		&\boldsymbol{x}_{k+1} = \boldsymbol{f}(\boldsymbol{x}_k,\boldsymbol{u}_k) + \boldsymbol{w}_k, \nonumber\\
		&\boldsymbol{y}_k \: \:\:\:\:= \boldsymbol{h}(\boldsymbol{x}_k,\boldsymbol{u}_k) + \boldsymbol{v}_k, \nonumber\\
		&\boldsymbol{g}(\boldsymbol{x}_k,\boldsymbol{u}_k) \leq 0, \quad k=0,\dots,N \label{subeq:constraints}
	\end{align}
\end{subequations}
where $N$ denotes the estimation horizon. The sequences ${X}_{0:n+1}$, ${W}_{0:n}, {V}_{0:n}$ represent matrices constructed from the data series $\boldsymbol{x}_k$, $\boldsymbol{w}_k$, and $\boldsymbol{v}_k$ respectively, such that ${X}_{0:n+1} = [\boldsymbol{x}_0, \boldsymbol{x}_1, ... , \boldsymbol{x}_{n+1}]$. The first term in the cost function, \eqref{subeq:cost_function}, minimizes the error between the initial predicted state and the prior initial state $\tilde{\boldsymbol{x}}_0$, and the second term penalizes the noise on the process and measurement model according to the corresponding weight matrices $P_{w}$ and $P_v$. The function~$\boldsymbol{g}(\cdot)$ in \eqref{subeq:constraints} embodies the state and input constraints.
\begin{figure*}[!t]
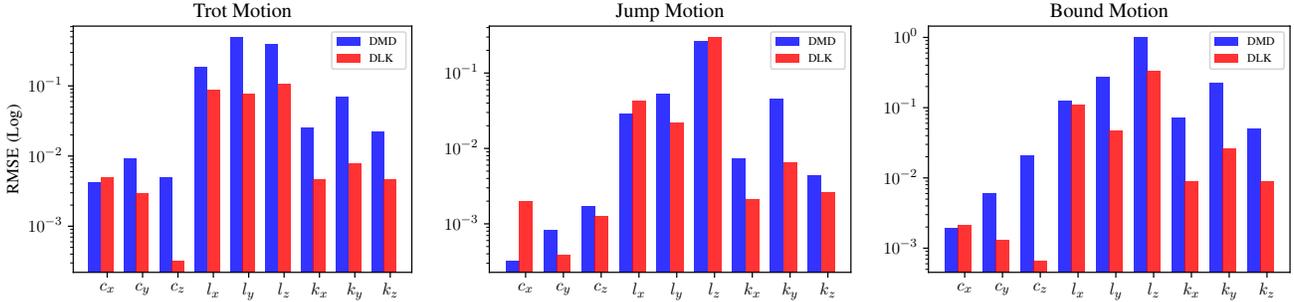

	\centering
	\begin{tabular}{ccc}
		\hspace{-5mm}\scalebox{0.58}{\input{files/rmse_trot.pgf}} & \hspace{-6mm} \scalebox{0.58}{\input{files/rmse_jump.pgf}} & \hspace{-6mm} \scalebox{0.58}{\input{files/rmse_bound.pgf}} \\
	\end{tabular}
	\vspace{-3mm}
	\caption{RMSE for open-loop simulation with learned linearized models on validation dataset. The deep learning Koopman model outperforms the dynamic mode decomposition model in terms of prediction accuracy nearly for all states across various motions.}
	\label{fig:rmse_comparison}
	\vspace{-6mm}
\end{figure*}

In the case of linear dynamics and linear measurement model, the constraints in \eqref{subeq:constraints} are replaced by the corresponding linear process and measurement models, as provided by the learned Koopman embedding, and $\boldsymbol{g}(\cdot)$ is replaced by proper bounds on state and input vectors. Consequently, the optimization problem becomes a convex QP due to the affine nature of the equality and inequality constraints.
\section{Results}
In this section, we detail the experimental framework employed to learn the Koopman embeddings for centroidal dynamics of the Unitree Go1 quadruped. Subsequently, we evaluate our learned models based on two key metrics: linearization error and their effectiveness in modeling different dynamic behaviors, specifically differentiating between fast and slow dynamic responses. This distinction is crucial for understanding the models' response in capturing the range of dynamic motions experienced by the robot. Further, we integrate the optimal linear dynamic model with the MHE framework to develop a data-driven centroidal state estimator. Lastly, we rigorously asses the performance of this estimator in comparison to the EKF over various dynamic motions of the quadruped robot.
\subsection{Experimental Setup}
Data were obtained at a sampling rate of $1$ kHz from a PyBullet simulation \cite{pybullet2021}, featuring Go1 quadruped executing a variety of dynamic gaits, generated by using a whole body motion planning framework \cite{avadesh2023}. The dataset for DLK comprised 10 hours of data capturing three distinct motions (trot, jump, and bound), each characterized by varying commanded base linear velocities. This dataset was divided such that 80\% was allocated for training purposes, while the remaining 20\% was reserved for validation and was not utilized during the training phase. Further, the DLK model was trained utilizing 8 hours of data from the training portion of the dataset. To construct DMD model, we selected 15 trotting and 15 jumping trajectories from the training set, each with a duration of 10 seconds and featuring diverse commanded base linear velocities.
\begin{table*}[!b]
	\centering
	\vspace{0mm}
	\caption{RMSE for centroidal state estimation with new noisy data not presented during training process. Our data-driven DMD-MHE estimator is outperforming model-based EKF in recovering the centroidal states across all dynamic motions.}
	\label{tab:rmse_values}
	\begin{tabular}{|p{1.4cm}|p{1.5cm}|p{1.5cm}|p{1.5cm}|p{0.8cm}|p{0.8cm}|p{0.8cm}|p{0.8cm}|p{0.8cm}|p{0.8cm}|p{0.8cm}|p{0.8cm}|p{0.8cm}|}
		\hline
		& \centering $c_x$ &\centering $c_y$ & \centering$c_z$	&\centering $l_x$ & \centering $l_y$ & \centering $l_z$ & \centering $k_x$ & \centering $k_y$ & $k_z$\\
		\hline
		\multicolumn{10}{|c|}{\textbf{Trotting Motion}} \\
		\hline
		\makecell{EKF\\DMD-MHE} &\makecell{$8\times10^{-4}$\\$\boldsymbol{6\times10^{-4}}$}& \makecell{$6\times10^{-4}$\\$\boldsymbol{2\times10^{-4}}$}& \makecell{$1\times10^{-4}$\\$\boldsymbol{6\times10^{-5}}$}& \makecell{$0.661$\\$\boldsymbol{0.040}$} & \makecell{$0.548$\\$\boldsymbol{0.034}$} & \makecell{$0.140$\\$\boldsymbol{0.048}$} &\makecell{$0.011$\\$\boldsymbol{0.003}$}&\makecell{$0.011$\\$\boldsymbol{0.005}$}& \makecell{$0.013$\\$\boldsymbol{0.005}$}\\
		\hline
		\multicolumn{10}{|c|}{\textbf{Jumping Motion}} \\
		\hline
		\makecell{EKF\\DMD-MHE}&\makecell{$1\times10^{-3}$\\$\boldsymbol{4\times10^{-4}}$}&\makecell{$1\times10^{-4}$\\$1\times10^{-4}$}& \makecell{$8\times10^{-4}$\\$8\times10^{-4}$}& \makecell{$0.230$\\$\boldsymbol{0.033}$} & \makecell{$0.161$\\$\boldsymbol{0.035}$} & \makecell{$0.643$\\$\boldsymbol{0.020}$} &\makecell{$0.021$\\$\boldsymbol{0.004}$}&\makecell{$0.055$\\$\boldsymbol{0.007}$}&\makecell{$0.022$\\$\boldsymbol{0.003}$} \\
		\hline
		\multicolumn{10}{|c|}{\textbf{Bounding Motion}} \\
		\hline
		\makecell{EKF\\DMD-MHE}&\makecell{$6\times10^{-4}$\\$\boldsymbol{4\times10^{-4}}$}&\makecell{$2\times10^{-4}$\\$\boldsymbol{1\times10^{-4}}$}&\makecell{$1\times10^{-3}$\\$1\times10^{-2}$} & \makecell{$0.283$\\$\boldsymbol{0.035}$} & \makecell{$0.173$\\$\boldsymbol{0.036}$} & \makecell{$0.498$\\$\boldsymbol{0.150}$} &\makecell{$0.026$\\$\boldsymbol{0.002}$}&\makecell{$0.280$\\$\boldsymbol{0.006}$}&\makecell{$0.030$\\$\boldsymbol{0.004}$} \\
		\hline
	\end{tabular}
	\vspace{-2mm}
\end{table*}
\subsection{Model Evaluation}
\begin{figure}[!t]
	\centering
	\scalebox{0.45}{\input{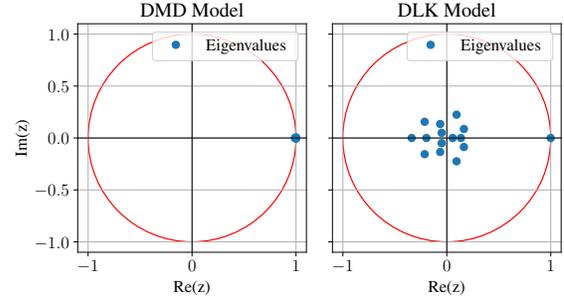}}
	\vspace{-1mm}
	\caption{Eigenvalues of the state-transition matrices for both learned linear models.}
	\label{fig:eigenvalues}
	\vspace{-4mm}
\end{figure}
In this study, we aimed to implement linear MHE that demands highly accurate predictive models to refine state estimation. The precision of the predictive model is especially critical for up to 50 steps ahead in moving horizon estimation, as longer prediction horizons can present practical challenges for real-time computation. We evaluated the performance of our models, DMD and DLK, focusing on their ability to accurately predict the centroidal states over an extended 500-step horizon. Figure \ref{fig:rmse_comparison} presents a summary of the RMSE values of the open-loop prediction of dynamics with our two models over the horizon of 500 steps for three various motions. Our results reveal that the DLK model outperforms the DMD model in terms of prediction accuracy nearly for all states. 

We also leverage the well-established control theories for linear dynamical systems to investigate the stability and dynamic characteristics of our linear models.
\begin{figure*}[!t]
	\center
	\scalebox{0.84}{\input{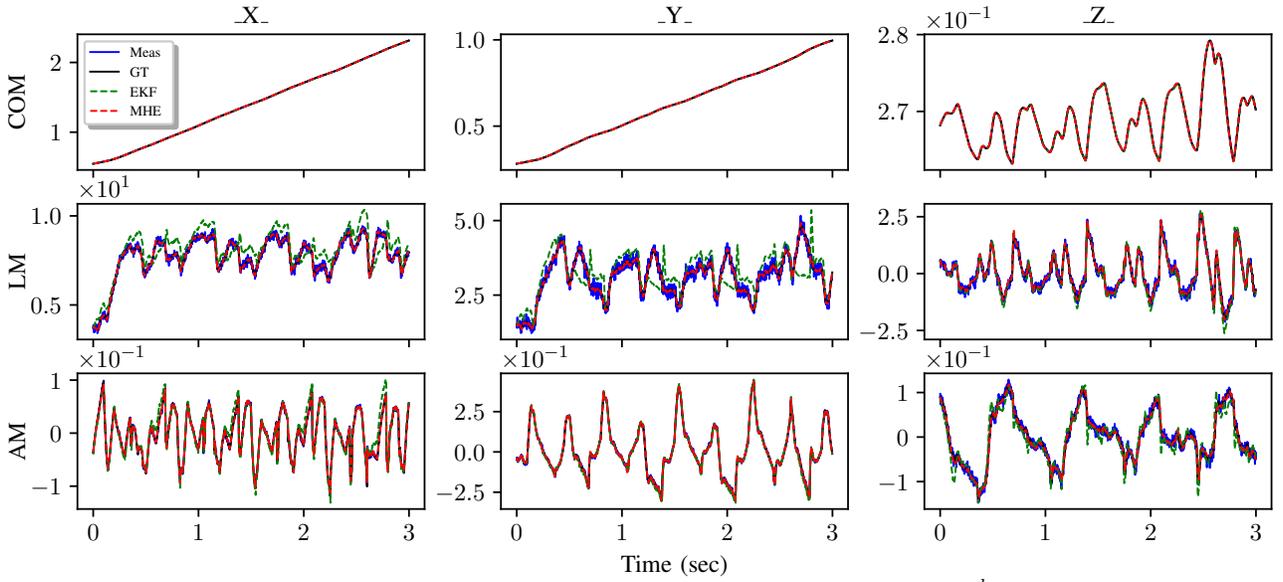}}
	\vspace{-3mm}
	\caption{Motion \#1: \textbf{Trot}. Estimation of the CoM position $(m)$ (top row), linear momentum $(\frac{kgm}{s})$ (middle row) and angular momentum $(\frac{kgm^2}{s})$ (bottom row), commanded base linear velocities during motion, $v_x =0.7 (\frac{m}{s})$ and $v_y =0.3 (\frac{m}{s})$. As evident, the intermittent rapid contact switching degrades the performance of EKF especially in estimating the linear momentum, while our estimator effectively maintains accurate state estimation during the motion.}
	\label{fig:trot}
	\vspace{-3mm}
\end{figure*}
\begin{figure*}[!t]
	\centering
	\scalebox{0.84}{\input{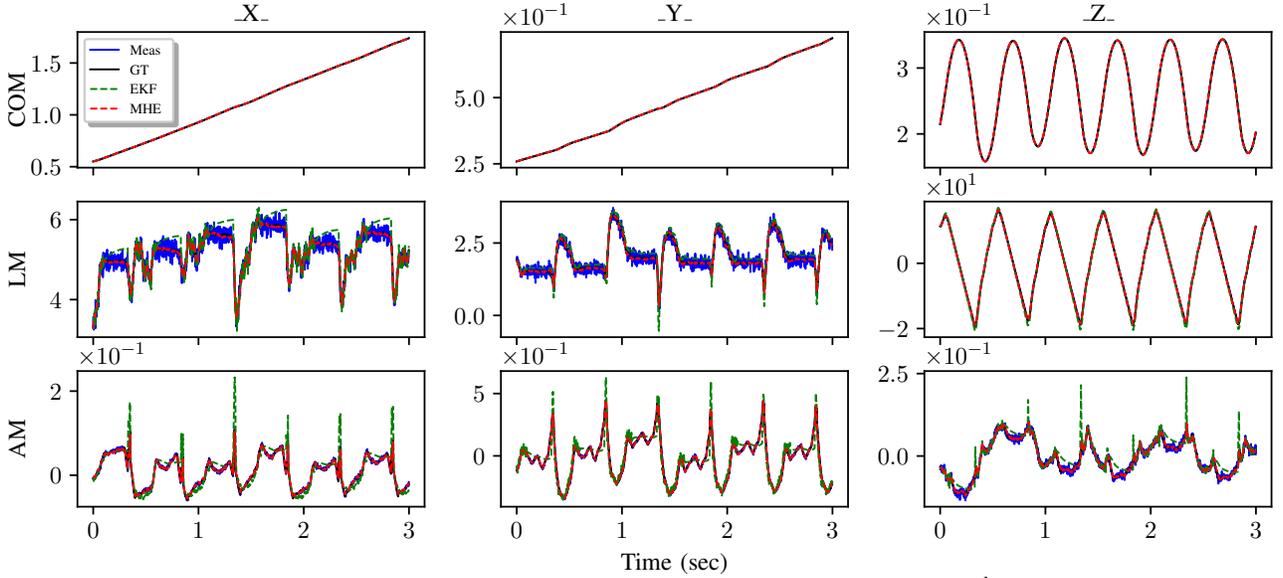}}
	\vspace{-3mm}
	\caption{Motion \#2: \textbf{Jump}. Estimation of the CoM position $(m)$ (top row), linear momentum $(\frac{kgm}{s})$ (middle row) and angular momentum $(\frac{kgm^2}{s})$ (bottom row), commanded base linear velocities during motion, $v_x =0.6 (\frac{m}{s})$ and $v_y =0.3 (\frac{m}{s})$. Large impacts at landing disrupts EKF by incorporating abrupt forces directly into the estimation process, while our estimator leverages a history of past measurements for precise and stable estimations.}
	\label{fig:jump}
	\vspace{-3mm}
\end{figure*}
To do so, we inspect the eigenvalues of the state-transition matrices for both models, revealing that the DLK model's eigenvalues are positioned closer to the unity circle's origin in the $Z$-plane than those of the DMD model, as illustrated in Fig. \ref{fig:eigenvalues}. Eigenvalues near the origin suggest a faster system response to inputs, which in this context, relate to the forces applied by the quadruped's end-effectors upon contact. This positioning leads to the DLK model having more precise predictions, albeit with a trade-off of a noisier dynamics representation compared to the DMD model. The role of the prediction model in state estimation is to mitigate measurement noise, making the DLK model's noisier dynamics unsuitable for direct state estimation applications. Therefore, we only utilized DMD model as our linear dynamics in MHE formulation, as it shows more robust dynamic response to noisy F/T sensory input.
\subsection{State Estimation}
To evaluate our estimator's performance, we introduced realistic simulated white Gaussian noise to the joint encoders (and propagated to joint velocities through numerical differentiation) and end-effector F/T sensor signals of the simulated Go1 model, replicating real-world sensor inaccuracies. This setup allowed for a comparison between our MHE approach and the conventional EKF \cite{rotella2015humanoid} across three dynamic motions, trotting, jumping, and bounding. These motions were selected for their complexity and the distinct challenges they present, including rapid state changes and sensor noise, and they have never been represented in the training dataset to ensure an unbiased evaluation. Ground truth data for the centroidal states were obtained from the simulation. Each motion lasted for $30 s$, with graphical representations focusing on the first $3 s$ to maintain clear and interpretable visuals. The Root Mean Squared Error (RMSE) metrics, detailed in Table \ref{tab:rmse_values}, quantitatively underscore our estimator's superiority in capturing the complex dynamics of quadruped locomotion with significantly enhanced precision of estimating centroidal state across all tested gaits.
\begin{figure*}[!t]
	\centering
	\scalebox{0.84}{\input{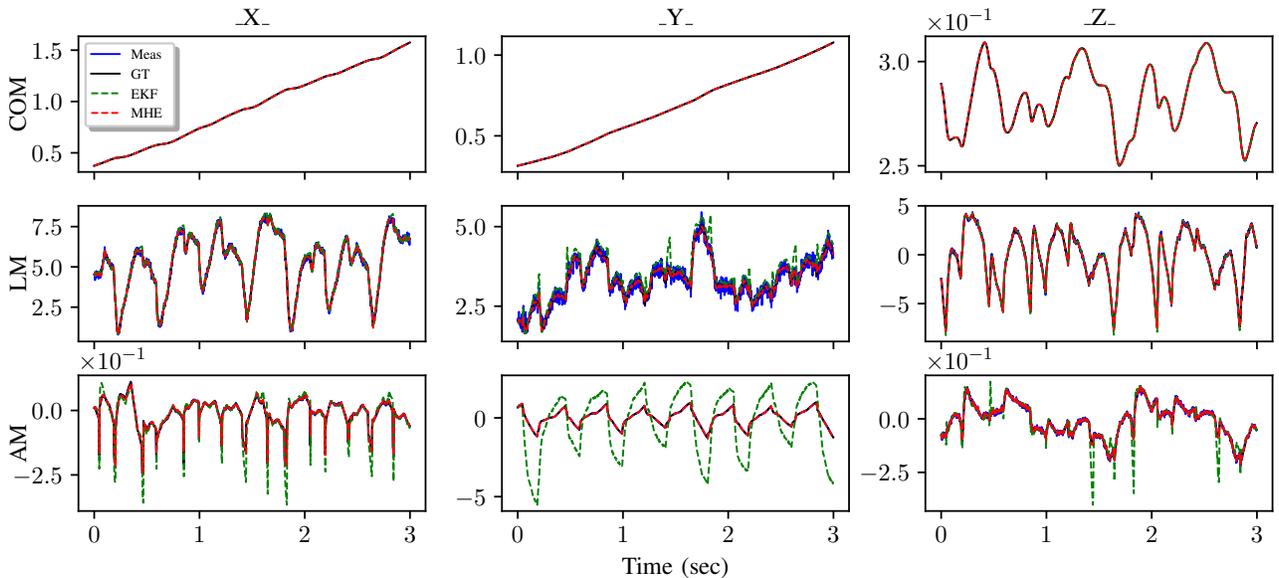}}
	\vspace{-3mm}
	\caption{Motion \#3: \textbf{Bound}. Estimation of the CoM position $(m)$ (top row), linear momentum $(\frac{kgm}{s})$ (middle row) and angular momentum $(\frac{kgm^2}{s})$ (bottom row), commanded base linear velocities during motion, $v_x = 0.6 (\frac{m}{s})$ and $v_y =0.3 (\frac{m}{s})$. This figure underscores our estimator's proficiency in accurately capturing the motion dynamics, especially the complex angular momentum changes in the $y$-axis, where EKF encounters estimation errors.}
	\label{fig:bound}
	\vspace{-5mm}
\end{figure*}
\subsubsection{Trotting Motion}
As shown in Fig. \ref{fig:trot}, this motion is characterized by intermittent rapid contact switching, posing significant challenges due to substantial noise in the F/T sensor readings. This noise notably impairs the EKF's ability to estimate linear momentum accurately. In contrast, our estimator consistently achieves precise state estimation throughout the motion.
\subsubsection{Jumping Motion}
The second motion presents significant challenges, especially due to the large impacts at landing that can disrupt EKF by incorporating abrupt forces directly into the estimation process potentially leading to estimation errors. Our method, which integrates a learned model into an MHE framework, counters these effects by drawing on a history of previous measurements. As shown in Fig. \ref{fig:jump}, our estimator maintains high accuracy and robustness, effectively filtering out noise from landing impacts and demonstrating a clear advantage in handling the complexities of jumping motion.
\subsubsection{Bounding Motion}
This motion's complexity is heightened by the rapid changes in angular momentum along the y-axis. Furthermore, the DMD model was not trained on this motion. However as depicted in Fig. \ref{fig:bound}, our estimator, by leveraging the accuracy of learned dynamics and a history of past measurements, achieves precise state estimation, effectively addressing the challenges presented by this complex motion. Conversely, EKF struggles to accurately capture this motion's nuances, resulting in errors in estimating angular momentum.
\section{Conclusions and Future Work}
In this work, we presented a novel centroidal state estimation approach for legged robots, critical for dynamic locomotion control. By leveraging the Koopman operator theory, we transformed the complex nonlinear dynamics of legged locomotion into linearized systems, using dynamic mode decomposition and deep learning. We evaluated the eigenvalues of both models and selected the most suitable model for estimation purposes within moving horizon estimator. Our data-driven estimator demonstrated superior performance over traditional Extended Kalman Filtering method in extensive simulations on a quadruped robot, showcasing robustness against the noise in force/torque sensor readings and accurately capturing the centroidal dynamics across various dynamic gaits.

In the future, we will implement our estimator on the physical Go1 Unitree robot and evaluate its performance in closed-loop control system, utilizing the estimated centroidal states. Furthermore, we plan to tackle the inherent challenges associated with dynamic model uncertainties by integrating online model identification within the estimation framework, to adaptively refine our models based on real-time performance. 

\bibliography{master} 

\begin{thebibliography}{10}

\bibitem{rotella2015humanoid}
N.~Rotella, A.~Herzog, S.~Schaal, and L.~Righetti, ``Humanoid momentum
  estimation using sensed contact wrenches,'' in {\em Proc.\ of the {IEEE/RAS}
  Int.\ Conf.\ on Humanoid Robots (Humanoids)}, 2015.

\bibitem{Koopman1931}
B.~O. Koopman, ``Hamiltonian systems and transformation in hilbert space,''
  {\em Proceedings of the National Academy of Sciences}, 1931.

\bibitem{Schmid2008dmd}
P.~Schmid and J.~Sesterhenn, ``Dynamic mode decomposition of numerical and
  experimental data,'' {\em Journal of Fluid Mechanics}, 2008.

\bibitem{Mezic2005}
I.~Mezi\'{c}, ``Spectral properties of dynamical systems, model reduction and
  decompositions,'' {\em Nonlinear Dynamics}, 2005.

\bibitem{Mezic2013}
I.~Mezi\'{c}, ``Analysis of fluid flows via spectral properties of the koopman
  operator,'' {\em Annual Review of Fluid Mechanics}, 2013.

\bibitem{bloesch2013state}
M.~Bloesch, M.~Hutter, M.~A. Hoepflinger, S.~Leutenegger, C.~Gehring, C.~D.
  Remy, and R.~Siegwart, ``State estimation for legged robots-consistent fusion
  of leg kinematics and imu,'' {\em Robotics}, 2013.

\bibitem{rotella2014state}
N.~Rotella, M.~Bloesch, L.~Righetti, and S.~Schaal, ``State estimation for a
  humanoid robot,'' in {\em Proc.\ of the {IEEE} Int.\ Conf.\ on Intelligent
  Robots \& Systems (IROS)}, 2014.

\bibitem{fallon2014drift}
M.~F. Fallon, M.~Antone, N.~Roy, and S.~Teller, ``Drift-free humanoid state
  estimation fusing kinematic, inertial and lidar sensing,'' in {\em Proc.\ of
  the {IEEE/RAS} Int.\ Conf.\ on Humanoid Robots (Humanoids)}, 2014.

\bibitem{camurri2020pronto}
M.~Camurri, M.~Ramezani, S.~Nobili, and M.~Fallon, ``Pronto: A multi-sensor
  state estimator for legged robots in real-world scenarios,'' {\em Frontiers
  in Robotics and AI}, 2020.

\bibitem{Dhedin2022visulainertial}
V.~Dh'edin, H.~Li, S.~Khorshidi, L.~Mack, A.~K.~C. Ravi, A.~Meduri, P.~Shah,
  F.~Grimminger, L.~Righetti, M.~Khadiv, and J.~Stueckler, ``Visual-inertial
  and leg odometry fusion for dynamic locomotion,'' {\em Proc.\ of the {IEEE}
  Int.\ Conf.\ on Robotics \& Automation (ICRA)}, 2022.

\bibitem{fourmy2021contact}
M.~Fourmy, T.~Flayols, P.-A. L{\'e}ziart, N.~Mansard, and J.~Sol{\`a},
  ``Contact forces preintegration for estimation in legged robotics using
  factor graphs,'' in {\em Proc.\ of the {IEEE} Int.\ Conf.\ on Robotics \&
  Automation (ICRA)}, 2021.

\bibitem{hartley2020contact}
R.~Hartley, M.~Ghaffari, R.~M. Eustice, and J.~W. Grizzle, ``Contact-aided
  invariant extended kalman filtering for robot state estimation,'' {\em Int.\
  Journal of Robotics Research (IJRR)}, 2020.

\bibitem{ramadoss2021}
P.~Ramadoss, G.~Romualdi, S.~Dafarra, F.~J.~A. Chavez, S.~Traversaro, and
  D.~Pucci, ``Diligent-kio: A proprioceptive base estimator for humanoid robots
  using extended kalman filtering on matrix lie groups,'' in {\em Proc.\ of the
  {IEEE} Int.\ Conf.\ on Robotics \& Automation (ICRA)}, 2021.

\bibitem{rotella2018contact}
N.~Rotella, S.~Schaal, and L.~Righetti, ``Unsupervised contact learning for
  humanoid estimation and control,'' in {\em Proc.\ of the {IEEE} Int.\ Conf.\
  on Robotics \& Automation (ICRA)}, 2018.

\bibitem{bledt2018contact}
G.~Bledt, P.~M. Wensing, S.~Ingersoll, and S.~Kim, ``Contact model fusion for
  event-based locomotion in unstructured terrains,'' in {\em Proc.\ of the
  {IEEE} Int.\ Conf.\ on Robotics \& Automation (ICRA)}, 2018.

\bibitem{geoff2020}
G.~Fink and C.~Semini, ``Proprioceptive sensor fusion for quadruped robot state
  estimation,'' in {\em Proc.\ of the {IEEE} Int.\ Conf.\ on Intelligent Robots
  \& Systems (IROS)}, 2020.

\bibitem{valsecchi2020quadrupedal}
G.~Valsecchi, R.~Grandia, and M.~Hutter, ``Quadrupedal locomotion on uneven
  terrain with sensorized feet,'' {\em IEEE Robotics and Automation Letters},
  2020.

\bibitem{grimminger2020open}
F.~Grimminger, A.~Meduri, M.~Khadiv, J.~Viereck, M.~W{\"u}thrich, M.~Naveau,
  V.~Berenz, S.~Heim, F.~Widmaier, T.~Flayols, {\em et~al.}, ``An open
  torque-controlled modular robot architecture for legged locomotion
  research,'' {\em IEEE Robotics and Automation Letters}, 2020.

\bibitem{Khorshidi2023}
S.~Khorshidi, A.~Gazar, N.~Rotella, M.~Naveau, L.~Righetti, M.~Bennewitz, and
  M.~Khadiv, ``On the use of torque measurement in centroidal state
  estimation,'' in {\em Proc.\ of the {IEEE} Int.\ Conf.\ on Robotics \&
  Automation (ICRA)}, 2023.

\bibitem{Bailly2021ddp}
F.~Bailly, J.~Carpentier, and P.~Souères, ``Optimal estimation of the
  centroidal dynamics of legged robots,'' in {\em Proc.\ of the {IEEE} Int.\
  Conf.\ on Robotics \& Automation (ICRA)}, 2021.

\bibitem{buchanan2021learning}
R.~Buchanan, M.~Camurri, F.~Dellaert, and M.~Fallon, ``Learning inertial
  odometry for dynamic legged robot state estimation,'' in {\em 5th Annual
  Conference on Robot Learning (CoRL)}, 2021.

\bibitem{Buchanan2023bias}
R.~Buchanan, V.~Agrawal, M.~Camurri, F.~Dellaert, and M.~Fallon, ``Deep imu
  bias inference for robust visual-inertial odometry with factor graphs,'' {\em
  IEEE Robotics and Automation Letters}, 2023.

\bibitem{Khadiv2023}
M.~Khadiv, A.~Meduri, H.~Zhu, L.~Righetti, and B.~Sch\"olkopf, ``Learning
  locomotion skills from mpc in sensor space,'' in {\em Proceedings of The 5th
  Annual Learning for Dynamics and Control Conference}, Proceedings of Machine
  Learning Research (PMLR), 2023.

\bibitem{Joshua2016dmdc}
J.~L. Proctor, S.~L. Brunton, and J.~N. Kutz, ``Dynamic mode decomposition with
  control,'' {\em SIAM Journal on Applied Dynamical Systems}, 2016.

\bibitem{Ian2019active}
I.~Abraham and T.~Murphey, ``Active learning of dynamics for data-driven
  control using koopman operators,'' {\em IEEE Transactions on Robotics}, 2019.

\bibitem{Williams2014edmd}
M.~Williams, I.~Kevrekidis, and C.~Rowley, ``A data-driven approximation of the
  koopman operator: Extending dynamic mode decomposition,'' {\em Journal of
  Nonlinear Science}, 2014.

\bibitem{lusch2018deep}
B.~Lusch, J.~N. Kutz, and S.~L. Brunton, ``Deep learning for universal linear
  embeddings of nonlinear dynamics,'' {\em Nature communications}, 2018.

\bibitem{takeishi2017learning}
N.~Takeishi, Y.~Kawahara, and T.~Yairi, ``Learning koopman invariant subspaces
  for dynamic mode decomposition,'' in {\em Advances in Neural Information
  Processing Systems}, 2017.

\bibitem{morton2018deep}
J.~Morton, A.~Jameson, M.~J. Kochenderfer, and F.~Witherden, ``Deep dynamical
  modeling and control of unsteady fluid flows,'' in {\em Advances in Neural
  Information Processing Systems}, 2018.

\bibitem{Bounou2021onlinekoopman}
O.~Bounou, J.~Ponce, and J.~Carpentier, ``Online learning and control of
  complex dynamical systems from sensory input,'' in {\em Advances in Neural
  Information Processing Systems}, 2021.

\bibitem{Wieber2006}
P.-B. Wieber, {\em Holonomy and Nonholonomy in the Dynamics of Articulated
  Motion}.
\newblock Springer Berlin Heidelberg, 2006.

\bibitem{vaswani2017attention}
A.~Vaswani, N.~Shazeer, N.~Parmar, J.~Uszkoreit, L.~Jones, A.~N. Gomez,
  {\L}.~Kaiser, and I.~Polosukhin, ``Attention is all you need,'' {\em Advances
  in neural information processing systems}, vol.~30, 2017.

\bibitem{rao2001constrained}
C.~V. Rao, J.~B. Rawlings, and J.~H. Lee, ``Constrained linear state
  estimation—a moving horizon approach,'' {\em Automatica}, 2001.

\bibitem{pybullet2021}
E.~Coumans and Y.~Bai, ``Pybullet, a python module for physics simulation for
  games, robotics and machine learning.'' GitHub repository, 2016--2021.

\bibitem{avadesh2023}
A.~Meduri, P.~Shah, J.~Viereck, M.~Khadiv, I.~Havoutis, and L.~Righetti,
  ``Biconmp: A nonlinear model predictive control framework for whole body
  motion planning,'' {\em IEEE Transactions on Robotics}, 2023.

\bibitem{franklin1998digital}
G.~F. Franklin, J.~D. Powell, M.~L. Workman, {\em et~al.}, {\em Digital control
  of dynamic systems}, vol.~3.
\newblock Addison-wesley Reading, MA, 1998.

\end{thebibliography}
\bibliographystyle{ieeetr}

\end{document}